\pdfoutput=1
\documentclass[11pt]{article}
\usepackage{graphicx}% so it makes black blobs
\usepackage[final]{acl}
\usepackage{enumitem}
\usepackage{times}
\usepackage{latexsym}

\usepackage[T1]{fontenc}
\usepackage[utf8]{inputenc}

\usepackage{microtype}

\usepackage{url}

\title{DocuT5: Seq2seq SQL Generation with Table Documentation}

\author{Elena Soare \\
  University of Glasgow \\
  elena.soare1998@gmail.com \\\And
  Iain Mackie \\
  University of Glasgow \\
  i.mackie.1@research.gla.ac.uk\\\And
  Jeffrey Dalton \\
  University of Glasgow \\
  jeff.dalton@glasgow.ac.uk
  }

\begin{document}
\maketitle
\begin{abstract}
Current SQL generators based on pre-trained language models struggle to answer complex questions requiring domain context or understanding fine-grained table structure. 
Humans would deal with these unknowns by reasoning over the documentation of the tables.  
Based on this hypothesis, we propose DocuT5, which uses off-the-shelf language model architecture and injects knowledge from external ``documentation'' to improve domain generalization. 
We perform experiments on the Spider family of datasets that contain complex questions that are cross-domain and multi-table. 
Specifically, we develop a new text-to-SQL failure taxonomy and find that 19.6\% of errors are due to foreign key mistakes, and 49.2\% are due to a lack of domain knowledge. 
We proposed DocuT5, a method that captures knowledge from (1) table structure context of foreign keys and (2) domain knowledge through contextualizing tables and columns.
Both types of knowledge improve over state-of-the-art T5 with constrained decoding on Spider, and domain knowledge produces state-of-the-art comparable effectiveness on Spider-DK and Spider-SYN datasets.  Our code is available at \url{https://github.com/grill-lab/DocuT5}.

\end{abstract}

\section{Introduction}

% --- Motivation ---
Translating natural language utterances into logical forms (SQL) that are executable against a relational database (text-to-SQL) is an important real-world task, reducing the barriers to entry for data analysis.
While leveraging pre-trained language models as SQL generators shows strong performance \cite{wang2020rat, lin2020bridging, scholak2021picard}, there are still limitations due to a lack of domain or table knowledge. 
This means that models struggle to adapt to new tables or domains, particularly when the natural language question does not explicitly reference the schema entities and relations.

% --- Problem ---

Cross-domain annotated benchmarks WikiSQL \cite{zhong2017wikisql} and Spider \cite{yu2018spider} contain a high percentage of natural language questions that directly reference column names contained within the database schema.
However, when column mentions are replaced with synonyms \cite{gan2021syn}, or paraphrases \cite{gan2021dk}, prediction accuracy scores drop substantially, even when questions containing domain knowledge are mentioned in the training data. 

% -- Related Work ---
Researchers leveraged graph neural networks which encoded the input as directed graphs \cite{bogin2019representing,  guo2019towards, wang2020rat}, augmented with syntactic metadata of the natural language question and database \cite{hui2022s2sql}, or enhanced the schema grounding modules \cite{liu2022semantic, wang2022proton}.
However, these are complicated and often dataset-specific architectures that are non-trivial to adapt to new domains or datasets.
On the other hand, Most seq2seq SQL generators \cite{hwang2019comprehensive, lin2020bridging, scholak2021picard} only linearize the database schema by enumerating table and column names, and lack the explicit table or domain knowledge for complex operations.

\begin{figure}[!h]
    \includegraphics[scale=0.2]{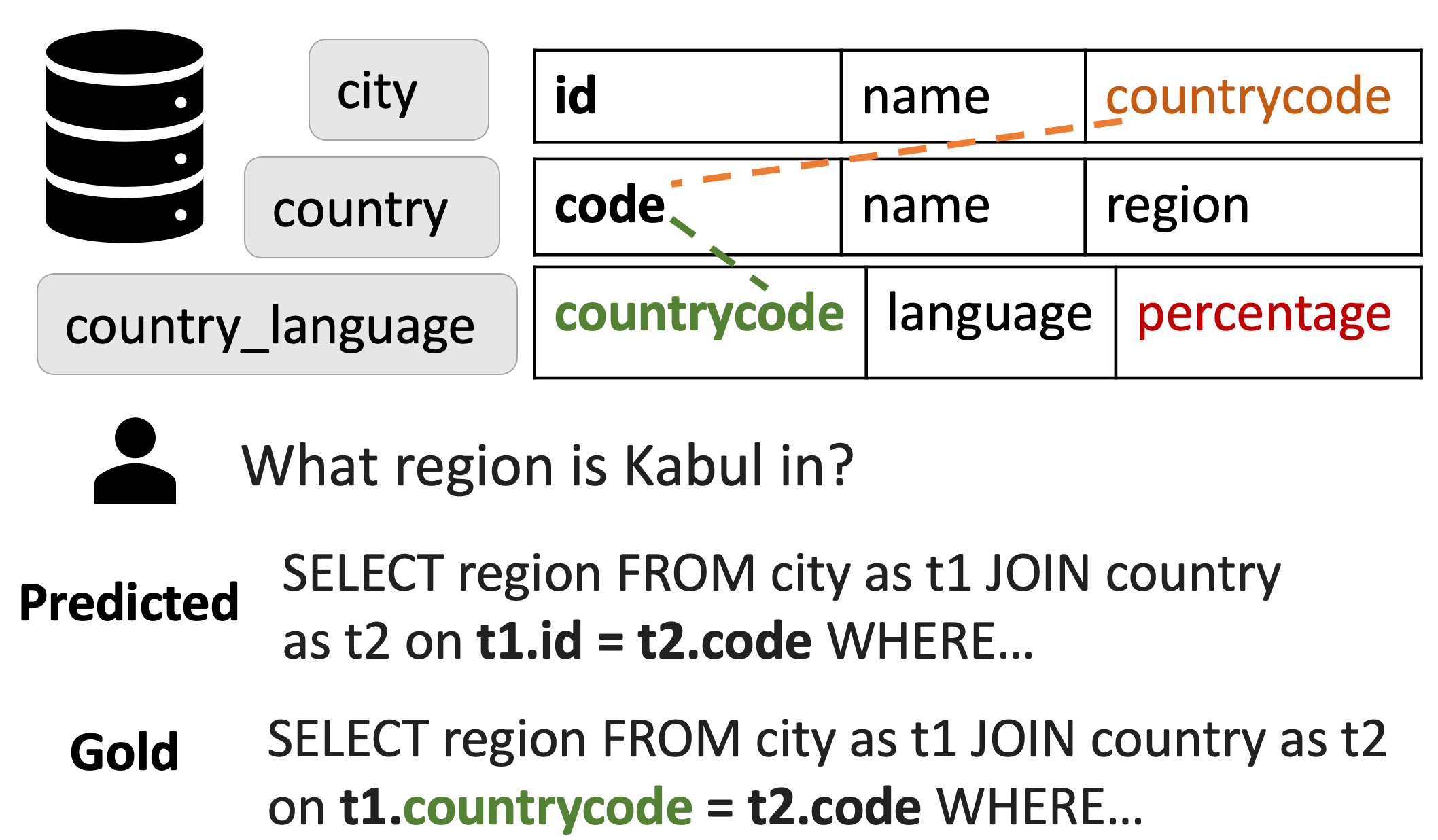}
    \caption{Foreign key failure example by seq2seq model \cite{scholak2021picard} on Spider dev.}
    \label{fig:incorrectfk}
\end{figure}
\vspace{-3mm}

\begin{figure}[!h]
    \includegraphics[scale=0.25]{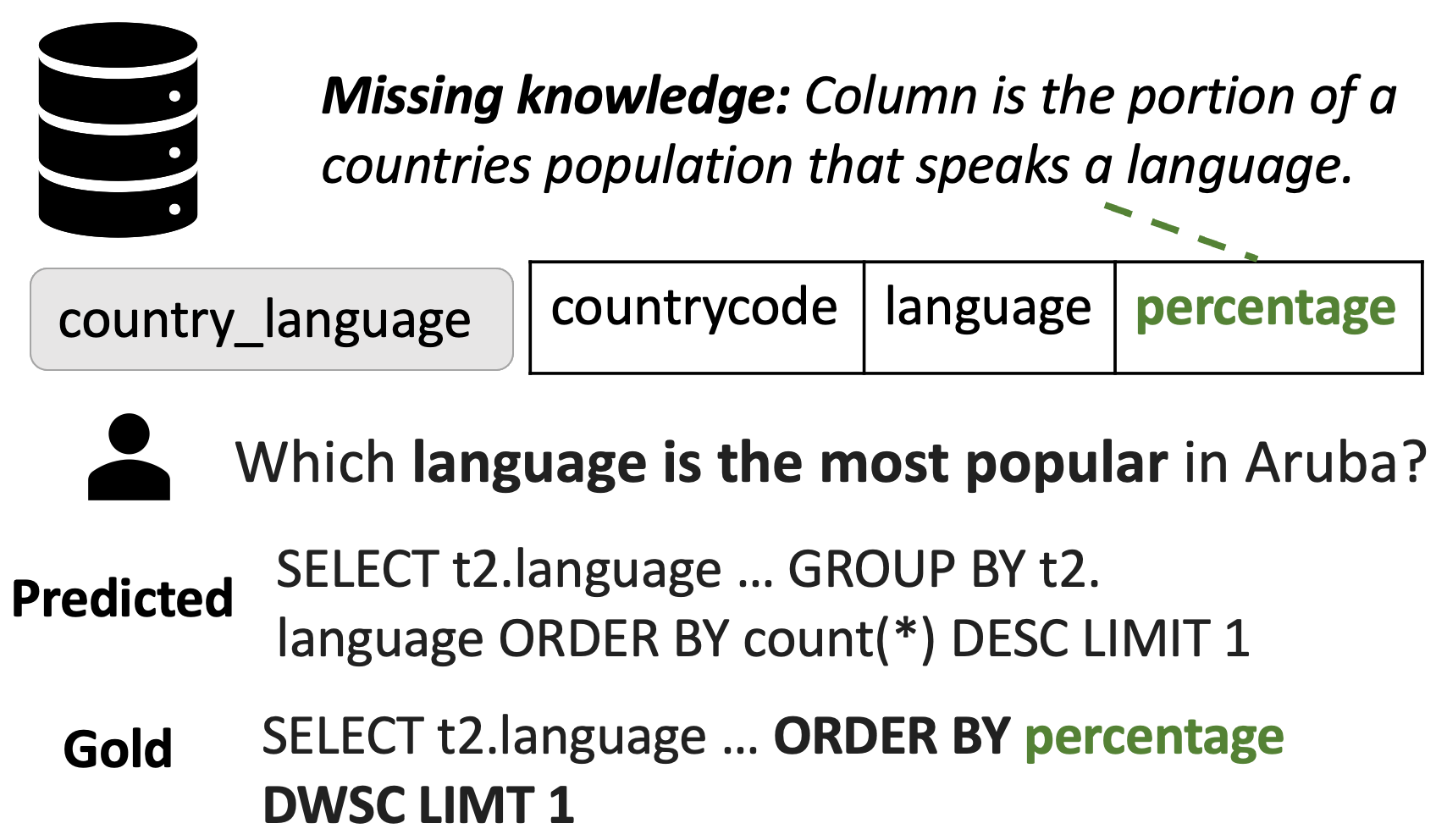}
    \caption{Domain knowledge failure example by seq2seq model \cite{scholak2021picard} on Spider dev.}
    \label{fig:incorrectdomain}
\end{figure}

% --- behavioral study ---
In this work, our first contribution is an in-depth behavioural study on a state-of-the-art seq2seq model with constrained decoding \cite{scholak2021picard}. 
We find that 19.6\% of errors are due to foreign key mismatches, and another 49.2\% are due to a lack of domain knowledge. 
For example, Figure \ref{fig:incorrectfk} shows a foreign key error where the model fails to identify \textit{countrycode} column in \textit{city} table is a foreign key to \textit{code} column in \textit{country} table.
Figure \ref{fig:incorrectdomain} shows the model fails to understand the table context that the \textit{percentage} column is the proportion of \textit{language} speakers in \textit{Aruba}.

\vspace{-3mm}

We argue that existing models suffer from these issues as they fail to incorporate knowledge in a way comparable to a human analyst. 
Specifically, when translating a request to a SQL query, a person would typically consult the table documentation, especially if it is a new domain or there is uncertainty about terminology.
Inspired by this approach, we propose DocuT5 that injects ``documentation'' into an off-the-shelf seq2seq model without requiring model modification, providing table and domain context during the generation process. 

Some recent work by \citet{dou2022unisar} has shown promising results in adding foreign key context into seq2seq models.
We take this a step further by proposing DocuT5, which adds rich domain knowledge via (1) a simplified means for encoding foreign keys (DocuT5-FK) and (2) textual schema descriptions that provide question context of the overall table and specific columns (DocuT5-SD). 
% We also combine encoding foreign keys and schema descriptions for the full DocuT5 model. 
Because the knowledge encoding is general, is it widely applicable to diverse language model classes and the results show consistent performance improvement across multiple dimensions.

% For our experiments, we adopt state-of-the-art seq2seq model PICARD \cite{scholak2021picard}, which leverages T5 \cite{raffel2019t5} a transformer-based model for text-to-text generation, trained on curated text with a masked language objective.
% We fine-tune T5 on Spider benchmark and evaluate it using the constrained decoder framework PICARD \cite{scholak2021picard}, which enforces well-formed SQL queries. 

% --- Results ---
We study the behavior of DocuT5 and previous models on the Spider family of datasets that contain complex questions that are cross-domain and multi-table.
This includes the original Spider dev set, as well as Spider-DK \cite{gan2021dk} that requires domain knowledge, and Spider-SYN \cite{gan2021syn} that replaces question terms with synonyms. 
We evaluate against comparable state-of-the-art SQL generators that leverage off-the-shelf pre-trained language models, including the BRIDGE, \cite{lin2020bridging}, PICARD \cite{scholak2021picard}, and UniSAr~\cite{dou2022unisar}.

Our experiments find that encoding foreign keys improves over T5 with constrained decoding by 3.1\% and comparable reported results by 2.2\% on Spider.
Separately, encoding schema description also improves on Spider against T5 with constrained decoding by 2.3\% and comparable reported results by 1.4\%.
Schema description also shows more significant relative improvements on Spider-DK (+2.4\%) and Spider-SYN (+2.8\%), where schema knowledge is most required.
Overall, DocuT5 achieves the best comparable results on Spider, Spider-DK, and Spider-SYN.
The contributions of this paper are:
\begin{itemize}
    \item \textbf{Spider behavior analysis}: develop a new text-to-SQL failure taxonomy and categories state-of-the-art seq2seq model failures on Spider, with 19.6\% attributed to foreign keys and 49.2\% attributed to domain knowledge.

    \item \textbf{Foreign keys}: demonstrate that encoding foreign key knowledge improves model performance by 3.1\% on Spider, achieving state-of-the-art against comparable models.  
    
    \item \textbf{Schema descriptions} Use schema schema descriptions to improves model performance by 2.3\% on Spider. 
    Incorporating domain knowledge produces the best reported results on Spider-DK and Spider-Syn datasets.
    % \item \textbf{DocuT5} Combing both forms of documentation ...
    
\end{itemize}

%%%%%%%%%%%%%%%%%%%%%%%%%%%%%%%%%%%%%%%%%%%%%%%%%%%%%%%%%%%%%%%%
%%%%%%%%%%%%%%%%%%%%%%%%%%%%%%%%%%%%%%%%%%%%%%%%%%%%%%%%%%%%%%%%

\section{Task}

We focus on the text-to-SQL generation task in a cross-domain setting. 
By definition, given a natural language question \textit{Q} and a database schema \textit{S}, we need to infer a correct \textit{SQL} query, which is executable against a database to retrieve an execution answer \textit{A} that satisfies \textit{Q}. 
Each database schema consists of a set of tables \textit{T}, and a set of columns \textit{C} belonging to the tables such that \textit{S = <T,C>}.
A foreign key \textit{FK} is a column which links two tables based on matching values and schema description \textit{SD} are metadata explaining \textit{S = <T,C>}. 

% The set of all table and column definitions, conveying the meaning of schema entities in relation to one another within a database, comprises the domain knowledge \textit{D} with respect to schema \textit{S}.

% \iain{I don't think this is our task? Our task is to generate \textit{Y} that correctly achieves \textit{A}...and we can use these components...}
% Our task is to encode \textit{Q}, \textit{S} and \textit{D} into a textual form that can be leveraged into a vanilla seq2seq model that generates correct SQL queries \textit{Y}, whose execution yields the correct execution answer \textit{A}.

\subsection{Datasets}

% We study the behavior of existing and new models on the standard Spider family of datasets.

\textbf{Spider} \cite{yu2018spider} is a cross-domain text-to-SQL dataset which contains 10,181 natural question utterances and 5,693 SQL statements for training, with a 535-question dev set. 
Spider contains 200 databases across a large variety of domains, and most queries contain at least one ORDER BY, JOIN, GROUP BY, or HAVING statements, unlike WikiSQL \cite{zhong2017wikisql}.
% The 200 databases within Spider encompass a variety of domains such as government, music, flights, academia, and geography. 
% Most queries contain at least one ORDER BY, JOIN, GROUP BY, or HAVING statements, unlike WikiSQL \cite{zhong2017wikisql}, which does not have any complex SQL statements. 
% Therefore, the models must learn diverse SQL compositional patterns while being mindful of domain knowledge.

\textbf{Spider-DK} \cite{gan2021dk} is a test set to assess the robustness of text-to-SQL models where questions require rarely observed domain knowledge or paraphrasing. 
% The authors demonstrate the prediction accuracy dramatically drops on samples that require domain knowledge, even if it appears in the training set.

\textbf{Spider-SYN} \cite{gan2021syn} is a variant developed by replacing schema-related words with synonyms that reflect real-world question paraphrases, eliminating explicit correspondence between questions and table schemas.
% The authors show that accuracy dramatically drops by eliminating explicit correspondence between natural language questions and table schemas, which are highly prevalent in the original Spider dev set.

\subsection{Metrics}

% We use standard text-to-SQL metrics to evaluate the accuracy of both the SQL and returned answer.

\textbf{Exact Set Match (EM)} \cite{yu2018spider} converts the predicted SQL query into an orderless dictionary, which are compared to the gold SQL queries based on string matching. 
Each clause has to be exact strings with the corresponding gold SQL clause for EM to be 1, otherwise EM is 0.  

\textbf{Execution Accuracy (EX)} is based on the results of executing SQL on the corresponding databases \cite{zhong2020executionaccuracy}. 
This metric deals with the semantic equivalence of SQL statements, which produce the same execution results despite being syntactically different.
Specifically, EX is 1 if the predicted and gold SQL have equivalent execution results, and 0 otherwise.

%%%%%%%%%%%%%%%%%%%%%%%%%%%%%%%%%%%%%%%%%%%%%%%%%%%%%%%%%%%%%%%%
%%%%%%%%%%%%%%%%%%%%%%%%%%%%%%%%%%%%%%%%%%%%%%%%%%%%%%%%%%%%%%%%

\section{Behavior Analysis} 
\label{section:picard}

In this section, we conduct an error analysis using a newly developed taxonomy to categorize errors within text-to-SQL models.
We reproduce T5 with constrained decoding \cite{scholak2021picard} on Spider dev set. 
We select the largest available T5 model (3 billion parameters) and activate the constrained decoding framework during inference. 
Based on the EX accuracy, we save the incorrectly inferred SQL queries from the Spider dev set. 
In total, there are 214 incorrect natural language questions and SQL pairs to manually categorise.

\subsection{Failure Taxonomy} 
\label{error-analysis}

The authors develop the following taxonomy of failure categories to identify text-to-SQL failure patterns.
It is important to understand ``how'' these models are failing to motivate future research directions, and this taxonomy can be used across any text-to-SQL dataset.
Experienced computer scientists with experience with SQL (the authors) discuss and agree on the categories of errors during annotation of the 214 failing Spider queries.
At least one author performs each annotation, and a majority vote resolves ambiguous instances. The failure taxonomy is:

\textbf{Incomplete Queries}
The predicted SQL queries are only partially decoded because the model predicts the ending tag prematurely.

\textbf{False Negatives}
The natural language question is ambiguous, or generated SQL queries are incorrectly labelled as inaccurate.

\textbf{Foreign Keys} errors are prevalent in JOIN operations where wrong foreign key columns are used. 
On a manual inspection, most of these errors are due to inconsistent naming within databases, making it non-trivial to reason over.

\textbf{Logical Errors} errors miss logical implications within the natural language question. 
For example, ordering people by age means ordering them in decreasing order of date of birth. 

\textbf{Domain Knowledge}
In this scenario, the model cannot capture the \textit{meaning} of the database schema due to missing table or column knowledge. 
We further divide this into subcategories:
\begin{itemize}[noitemsep,nolistsep]
    \item \textbf{Aggregation Errors}: Incorrect use of SUM, COUNT, MIN, and MAX functions.
    \item \textbf{Incorrect Table}: DB entries are retrieved from the incorrect table.
    \item \textbf{Incorrect Column}: Predicting the wrong columns in SELECT or WHERE clauses. Often attributable to implicitly referenced columns within the question. 
    \item \textbf{Value Errors}: Incorrect values in WHERE clauses. Values mentioned in questions are not exact strings matches to the DB entries.
    \item \textbf{Complex Errors}: The model fails to understand a mix of paraphrases, complex table structures, and difficult required aggregations.
\end{itemize}

\subsection{Findings} 

Table \ref{picard-error} shows the breakdown of T5 with constrained decoding errors on the Spider dev set.
We find that 22.4\% of questions are False Negative, which is not a suitable target category for model improvement.
Furthermore, 6.2\% of errors are due to Incomplete SQL and 2.8\% are due to Logical Errors, both relatively small failure categories.

We specifically focus on Foreign Keys and Domain Knowledge categories based on the disproportionate amount of errors.
For example, 19.6\% of errors are due to table joins from incorrect foreign keys. 
The overall errors caused by Domain Knowledge totalled 49.2\% across five subcategories, with Aggregation Errors (17.2\%), Incorrect Column (13.4\%), and Complex Errors (11.0\%). 

\begin{table}[h!]
\centering
\begin{tabular}{|l|l|}
    \hline
    \multicolumn{1}{c}{\bfseries Failure Categories} & \multicolumn{1}{c}{\bfseries Percentage}  \\ \hline
    Incomplete SQL & 6.2\% \\ \hline
    False Negatives & 22.4\% \\ \hline
    Foreign Keys & 19.6\% \\ \hline
    Logical Errors & 2.8\% \\ \hline
    DK - Incorrect AGG & 17.2\% \\ \hline
    DK - Incorrect Table & 3.8\%\\ \hline
    DK - Incorrect Column & 13.4\%\\ \hline
    DK - Incorrect Value & 3.8\%\\ \hline
    DK - Complex & 11.0\%\\ \hline
\end{tabular}
\caption{Error Analysis on Spider dev based on T5+3B with constrained decoding.}
\label{picard-error}
\end{table}

We investigate whether giving models access to contextual ``documentation'' helps with Foreign Keys and Domain Knowledge failure categories.
Therefore, this categorization is important because it helps us assess what context will help the model for specific failure categories.
For example, Value Errors motivate contextualizing column values, especially for hard-to-interpret boolean columns. 
Incorrect Column errors are due to the model not mapping question intent with the correct column, requiring more context to ground the column for specific use cases. 
Lastly, Incorrect Aggregation is when the model is having issues discerning the value type of a specific column and explicitly stating whether the column is a date or number would be beneficial.

%%%%%%%%%%%%%%%%%%%%%%%%%%%%%%%%%%%%%%%%%%%%%%%%%%%%%%%%%%%%%%%%
%%%%%%%%%%%%%%%%%%%%%%%%%%%%%%%%%%%%%%%%%%%%%%%%%%%%%%%%%%%%%%%%

\section{Method}

DocuT5 is a new method to include documentation into seq2seq models that is general and widely applicable to diverse models without requiring changes to model structure.
Specifically, we focus on improving schema serialization through explicit foreign keys mapping and domain knowledge through additional table and column context. 
As is standard for seq2seq models, we concatenate the natural language question and serialized database schema and ask the model to generate the SQL.

We employ an off-the-shelf pre-trained language model T5 \cite{raffel2019t5}, experimenting with both the T5-Base and T5-Large variants. 
For all DocuT5 variants, we serialized the database schema by enumerating the table and column names, similar to \cite{scholak2021picard}.
We also encode database content snippets (anchor texts) by performing fuzzy string matching on the natural language question and the database entries \cite{scholak2021picard, lin2020bridging}.
% , representing anchor texts are placed between the special tokens '(' and ')'.
Lastly, to reduce non-executable hallucinations, we employ constrained decoding to incrementally perform sanity checks at inference \cite{scholak2021picard}.

% We propose one method to inject schema knowledge using Foreign Keys (FORKT5) and one method to inject schema-specific domain-knowledge by augmenting the input with schema descriptions (DORKT5). 
% We employ the pre-trained language model T5 \cite{raffel2019t5} in all our experiments, whose input consists of a concatenation of the natural language question and serialization of the database schema.

\subsection{DocuT5-FK: Foreign Keys} 
\label{fk}

\begin{figure}[h!]
\begin{center}
    \includegraphics[scale=0.3]{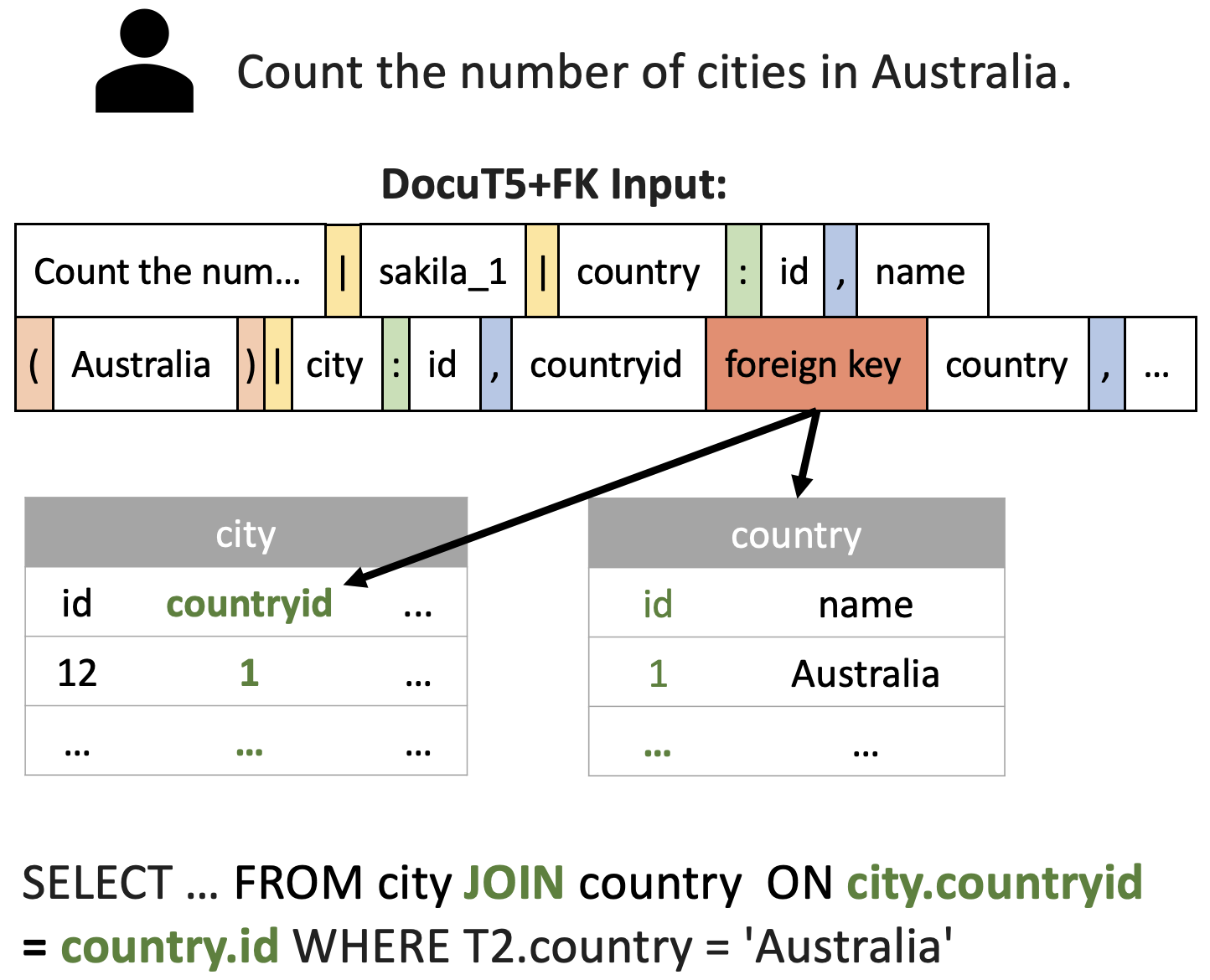}
    \end{center}
    \caption{DocuT5-FK: jointly encoding the natural language question and database schema, including foreign key relations. After every foreign key column, we add a special marker ``foreign key'', followed by the referenced table name.}
\label{archfk}
\end{figure}
\vspace{-3mm}

Encoding cross-table relations in textual form is more challenging than, for instance, column-table relations (column belongs to table) because foreign key relations are more intuitively mapped to directed graphs.
To reduce complexity and exploit the ability of pre-trained language models to grasp free-form text better than structured data, we add a special ``foreign key'' marker after every foreign key column in the serialized schema representation, followed by the name of the table it is referenced to. 
For example, Figure \ref{archfk} shows the input design for the \textit{city} table, where the foreign key \textit{column} is a reference to the \textit{id} in table \textit{country}.
We add only the table reference next to FK column \textit{country}: ``foreign key country''.

% We do not add the column name reference as well, because we aim to inject some knowledge of cross-table relations, rather than creating a complex flattened schema representation whose compositional patterns can be difficult to learn by a model trained solely on unstructured text.

% \elena{not sure about generalization, but in general simple patterns are easier to be picked up by the model?}
% \elena{less complex than for instance UniSar, who encode foreign keys including the tables and the respective FK column, we are just including the table the FK is referencing to}

\begin{center}
\begin{figure*}[h!]
\centering
\includegraphics[scale=0.29]{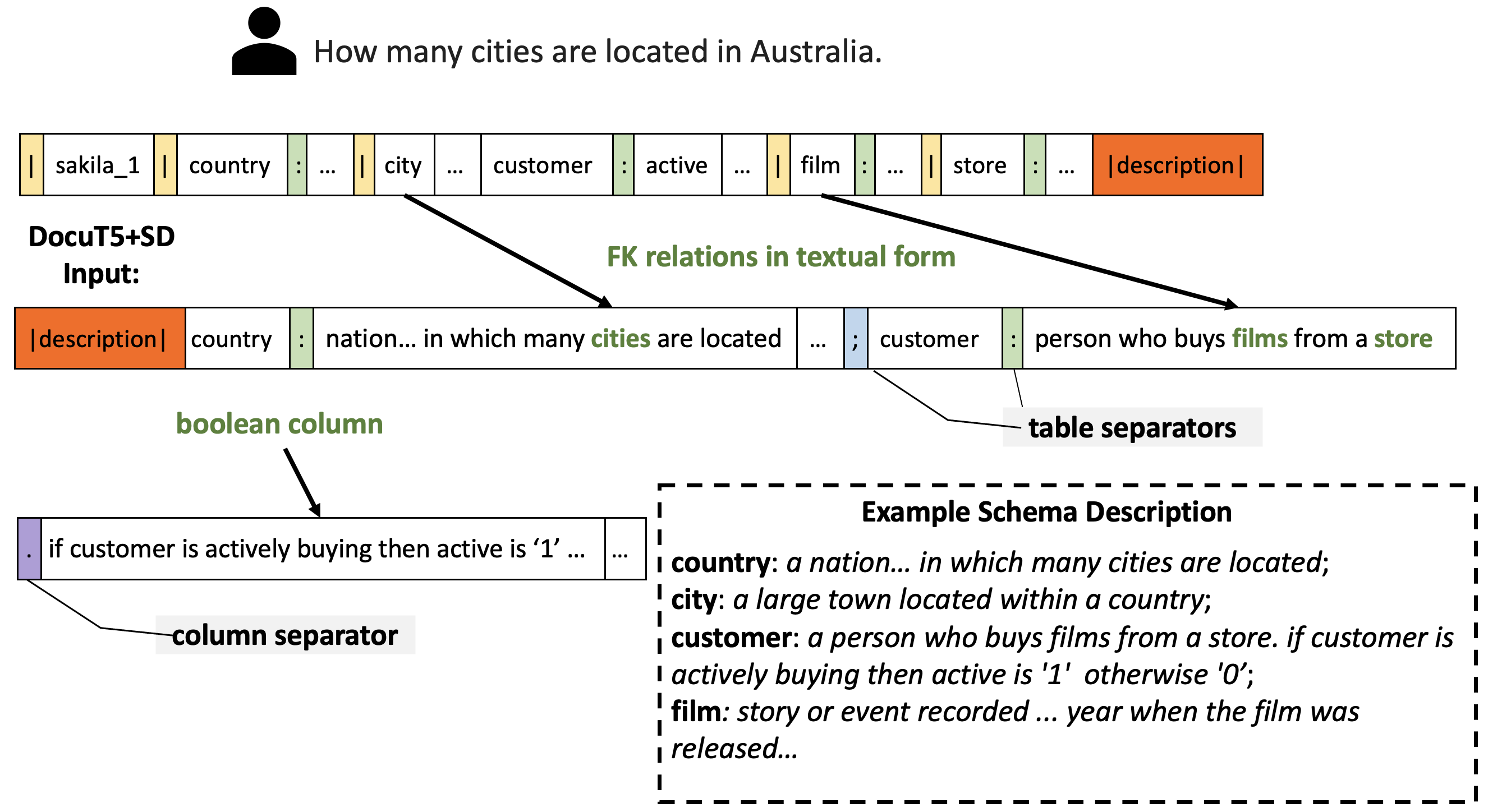}
\caption{DocuT5-SD: Adding semi-structured schema descriptions, which encapsulate schema-specific domain knowledge for both tables and columns.}
\label{fig:tableaug-arch}
\end{figure*}
\end{center}
\vspace{-3mm}

Based on initial experimentation, we find that simple patterns for encoding foreign key information are easier for seq2seq models to learn. 
Therefore, we avoided more complex column-to-column foreign keys , as used in UniSAr, including only the table name and linking the foreign key table directly in the schema serialization.

We also only selectively add the most beneficial foreign key relation types to keep input less complex. Specifically:

\textbf{1-to-1 relations} associate one record in a table with another record in another table. 
We add the FK relation marker after \textit{both} columns if they represent primary keys,
otherwise, we add the relation marker \textit{only} after the non-primary key column.

\textbf{1-to-Many relations} associate one record in a table with multiple records in another table. 
In this case, we add the foreign key relation \textit{only} after the foreign key column on the \textit{-Many} sides.

\subsection{DocuT5-SD: Schema Descriptions}

As identified in Section \ref{error-analysis}, around 50\% of current seq2seq errors result from a lack of domain knowledge. 
Specifically, the model makes errors due to missing table or column knowledge.
For example, when asked \textit{Which language is the most popular in Aruba}, current models struggle to identify \textit{percentage} as the target column and what logical operations are required to compute the answer.

% Injecting domain knowledge in the external memory of our model through input augmentation allows an easier revision and constraint of data quality. 
% The goal is to inject domain knowledge, pinpointed to schema entities, while keeping descriptions structured to inject compositional bias.

To solve for missing domain knowledge, we introduce schema descriptions directly as textual input to leverage pre-trained language models' ability to reason over text.  
These descriptions aim to provide paraphrases or database context to create a more specific table definitions. 
For example, a column named \textit{date} within table \textit{company} can mean the date that the company was founded or when the database table was last updated, depending on the contents. 
Furthermore, we tried to inject cross-table relations into natural language by describing schema entities concerning other entities. 
Figure \ref{fig:tableaug-arch} shows the model input that explains the table \textit{customer}, in relation to table \textit{store} and \textit{film}.

Computer scientists with working knowledge of SQL and access to a commercial search engine (the authors) develop these high-quality schema descriptions, which will be released as part of the paper.
Figure \ref{fig:tableaug-arch} shows how the schema descriptions are included alongside the question and serialized schema, separated by the ``description'' marker.
Annotators follow the subsequent guidelines to keep descriptions short, specific and consistent: 

%are written for all tables within a database schema
\textbf{Table definitions} describe all tables within a database schema.  

\textbf{Column definitions} describe only ambiguous columns, i.e. name is a partial match to its natural language reference (\textit{indepyear} is explained as \textit{independence year}) or vague. 

\textbf{Boolean columns} make implicit use of domain knowledge and are often ambiguous for pre-trained language models \cite{gan2021dk}. 
In Figure \ref{fig:tableaug-arch}, the definition of column \textit{active} in the DB \textit{sakila\_1} contains explanations of the values ``0'' and ``1''.

\textbf{Structured descriptions}: We keep the table description semi-structured by ordering the table and column paraphrases with the same success as the baseline serialized schema, with different separator tokens between table and columns.

% \subsection{DocuT5: Serialization and Descriptions}

% The full DocuT5 model combines both the encoded schema serialisation and schema descriptions outlined in the prior two subsections.
% Specifically, this approach seeks to leverage documentation of both the foreign keys, table definitions, and column descriptions. 

%%%%%%%%%%%%%%%%%%%%%%%%%%%%%%%%%%%%%%%%%%%%%%%%%%%%%%%%%%%%%%%%
%%%%%%%%%%%%%%%%%%%%%%%%%%%%%%%%%%%%%%%%%%%%%%%%%%%%%%%%%%%%%%%%

\section{Experiments and Evaluation}

In this section we report the experimental results for including documentation into seq2seq models. 
% We evaluate the following hypotheses:

% \begin{enumerate}[label={\textbf{H\arabic*}}, noitemsep]
%   \item \label{th:H1} Schema encoding with foreign keys improves overall prediction accuracy, especially helping with SQL queries of the Foreign Keys failure category.
%   \item \label{th:H2} Schema descriptions will provide useful question context and improve overall prediction accuracy, especially helping with SQL queries of the Domain Knowledge failure categories.
% %   \item \label{th:H3} Combining encoding of foreign keys and schema descriptions will improve overall prediction accuracy.
% \end{enumerate}

%%%%%%%%%%%%%%%%%%%%%%%%%%%%%%%%%%%%%%%%%%

\subsection{Experimental Setup}

To evaluate our methods, we implement comparable and strong baseline SQL generation systems that use off-the-self pre-trained language models. 
Specifically, we fine-tune \textbf{T5-Base} and \textbf{T5-Large} models on the Spider train set, with evaluation on the Spider dev set \cite{scholak2021picard}.
We use the standard Huggingface checkpoints with cross-entropy loss, a constant learning rate of 0.0001 with an Adafactor optimizer, and train for around 400 epochs. 
We perform an early stop and select the best scoring model by EM on Spider dev set, and evaluate zero-shot on Spider-DK and Spider-SYN  testsets.
Additionally, we evaluate performance of both \textbf{T5-Base\textsuperscript{CD}} and \textbf{T5-Large\textsuperscript{CD}} with constrained decoding \cite{scholak2021picard}.

We also compare against the strongest reported results from comparable pre-trained language model SQL generators. 
% \textbf{IRNet}  \cite{guo2019towards} performs a schema linking and synthesize a SemQL query and use domain knowledge.
% \textbf{RAT-SQL} \cite{wang2020rat} utilizes relation-aware self-attention mechanism to abstract various pre-defined relations, and \textbf{RAT-SQL+Bert} \cite{wang2020rat} extends this by include BERT-based representations.
\textbf{PICARD} \cite{scholak2021picard} uses constrained decoding and we compare against the comparable T5-Base and T5-Large models.
\textbf{BRIDGE} \cite{lin2020bridging} uses a BERT-large model for contextualization and a pointer-generator decoder.
The recently published \textbf{UniSAr} \cite{dou2022unisar} encodes table structures within the textual input to BART-Large and uses constrained decoding.

Similar to how we employ the T5 models, we train \textbf{DocuT5-FK} and \textbf{DocuT5-SD} using T5-Base and T5-Large models under the same training and evaluation methodology.
We also evaluate performance of \textbf{DocuT5-FK\textsuperscript{CD}} and  \textbf{DocuT5-SD\textsuperscript{CD}} with constrained decoding \cite{scholak2021picard}. 

\subsection{Results}

% ou: Examples I found interesting: https://docs.google.com/document/d/1QhoaUb3-kCRxU00UaB0wBW-QhNxdcDJNZ-qTEHxfsM4/edit?usp=sharing

We provide the results and analysis across schema serialization and schema descriptions. 

\subsubsection{Foreign Keys}

\begin{center}
\begin{table}[h!]\renewcommand{\arraystretch}{1}
\centering
\begin{tabular}{c c cc c }
    \hline
    \multicolumn{2}{c}{} &
    \multicolumn{2}{c}{Spider dev}\\
    \cline {3-4} 

    \multicolumn{1}{c}{Model}  && EM & EX \\
    \hline
    % \multicolumn{1}{|l|}{IRNet}     && 53.2\%  & -   \\
    % \multicolumn{1}{|l|}{RAT-SQL}       && 62.7\%  & - \\
    % \multicolumn{1}{|l|}{RAT-SQL+Bert}     && 69.7\% & - \\
    \multicolumn{1}{|l|}{PICARD-Base}     && 65.8\% & 68.4\%  \\
    \multicolumn{1}{|l|}{PICARD-Large}     && 69.1\% & 72.9\%  \\
    \multicolumn{1}{|l|}{BRIDGE}   && 70.0\% & -   \\
    \multicolumn{1}{|l|}{UniSAr}      && 70.0\%  &  \\
    \hline
    \multicolumn{1}{|l|}{T5-Base}                            && 59.4\% & 60.0\% \\
    \multicolumn{1}{|l|}{DocuT5-Base-FK}                     && 57.1\% & 60.1\% \\ 
    \multicolumn{1}{|l|}{T5-Base\textsuperscript{CD}}        && 66.6\% & 68.4\% \\
    \multicolumn{1}{|l|}{DocuT5-Base-FK\textsuperscript{CD}} && 67.0\% & 72.0\% \\ 
    \hline
    \multicolumn{1}{|l|}{T5-Large}                          && 65.3\% & 67.2\% \\
    \multicolumn{1}{|l|}{DocuT5-Large-FK}                   && 64.3\% & 67.5\% \\ 
    \multicolumn{1}{|l|}{T5-Large\textsuperscript{CD}}      && 69.1\% & 72.9\% \\ 
    \multicolumn{1}{|l|}{DocuT5-Large-FK\textsuperscript{CD}} && \textbf{72.2\%} & \textbf{77.0\%} \\ 
    \hline
  \end{tabular}
 \caption{Spider results: serializing schema with foreign keys. Our results (bottom and middle) and relevant prior work (top). (\textsuperscript{CD}) highlights when using constrained decoding \cite{scholak2021picard}. Prior work: PICARD  \cite{scholak2021picard}, BRIDGE \cite{lin2020bridging}, and UniSAr \cite{dou2022unisar}}
 \label{table:results-fk} 
 \end{table}
\end{center}

% --- Results ---
% Unsurprisingly, constrained decoding helped across all datasets for both Base and Large model variants. 
% Foreign keys also help within Spider-DK, improving EM by 1.1\% and EX by 1.7\%.
% However, EM is marginally worse on Spider-SYN by 0.4\% but 1.7\% better on EX.
% It is particularly interesting to compare our results to UniSar, who sought to included foreign key information in the text input via a different method.
\vspace{-3mm}
Table \ref{table:results-fk} shows the schema serialization results where foreign keys are explicitly encoded.
DocuT5-Base-FK\textsuperscript{CD} improves over the comparable T5-Base\textsuperscript{CD} model by 0.4\% on EM and 1.6\% on EX on Spider dev.
While results are even more impressive when focusing on T5-Large models.
DocuT5-Large-FK\textsuperscript{CD} improves over T5-Large\textsuperscript{CD} by 3.1\% on EM and 3.9\% on EX on Spider dev. 
DocuT5-Large-FK\textsuperscript{CD} is also at least 1.4\% better than reported results for similar-sized models, such as UniSar.
This is strong evidence that our simple method of injecting foreign key knowledge is highly effective.

\begin{center}
\begin{table*}[h!]\renewcommand{\arraystretch}{1}
\centering
\begin{tabular}{c c cc c cc c cc }
    \hline
    \multicolumn{2}{c}{} &
    \multicolumn{2}{c}{Spider dev} & &\multicolumn{2}{c}{Spider-DK} && \multicolumn{2}{c}{Spider-SYN}\\
    \cline {3-4} 
    \cline {6-7} 
    \cline {9-10}  
    
    \multicolumn{1}{c}{Model}  && EM & EX && EM & EX && EM & EX\\
    \hline
    % \multicolumn{1}{|r|}{T5-Large\cite{scholak2021picard}} & t5.1.1 && 71.2\% & 74.4\% && TBA\% & TBA\% && TBA\% & TBA\% \\
    % \multicolumn{1}{|r|}{T5-Large+SD}&t5.1.1 && 66.44\% & 71.37\% && 44.11\% & 54.77\% && 52.9\% & 57.83\%  \\
    % \multicolumn{1}{|r|}{T5-Large+PICARD \cite{scholak2021picard}}& t5.1.1 & & 74.8\% & 79.2\% && 51.4\% & 59.81\% && 60.83\% & 66.83\% \\ 
    % \multicolumn{1}{|r|}{T5-Base+SD+PICARD}& t5.1.1 && 73.3\% & 78.8\% && 48.04\% & 61.5\% && 61.9\% & 68.18\% \\ \cline{1-2} \cline{4-5}
    % \multicolumn{1}{|l|}{IRNet  \cite{guo2019towards}}     && 53.2\%  & -  && 33.1\% & - && 28.4\% & - \\
    % \multicolumn{1}{|l|}{RAT-SQL \cite{wang2020rat}}       && 62.7\%  & -  && 35.8\% & - && 33.6\% & - \\
    % \multicolumn{1}{|l|}{RAT-SQL+B \cite{wang2020rat}}     && 69.7\% & -  && 40.9\% & - &&  48.2\% & - \\
    \multicolumn{1}{|l|}{PICARD-B\cite{scholak2021picard}}     && 65.8\% & 68.4\%  && - & - && - & - \\
    \multicolumn{1}{|l|}{PICARD-L \cite{scholak2021picard}}     && 69.1\% & 72.9\%  && - & - && - & - \\

    \multicolumn{1}{|l|}{BRIDGE \cite{lin2020bridging}}   && 70.0\% & -  && - & - && - & - \\
    \multicolumn{1}{|l|}{UniSAr \cite{dou2022unisar}}      && 70.0\%  & -  && - & - && - & - \\
    \hline
    \multicolumn{1}{|l|}{T5-Large}            && 65.3\% & 67.2\%  && 39.8\% & 46.5\% && 53.4\% & 57.3\% \\
    \multicolumn{1}{|l|}{DocuT5-Large-SD}           && 64.4\% & 66.3\% && 41.9\% & 50.5\% && 52.2\% & 55.7\% \\
    \multicolumn{1}{|l|}{T5-Large\textsuperscript{CD}}  && 69.1\% & 72.9\% && 45.6\% & 55.0\% && 58.9\% & 64.5\% \\
    \multicolumn{1}{|l|}{DocuT5-Large-SD\textsuperscript{CD}} && \textbf{71.4\%} & \textbf{74.7\%} && \textbf{48.0\%} & \textbf{59.6\%} && \textbf{61.7\%} & \textbf{68.2\%} \\ \hline
  \end{tabular}
 \caption{Spider Results: Schema Description. Our results (bottom) and relevant prior work (top). (\textsuperscript{CD}) highlights when using constrained decoding \cite{scholak2021picard}.}
  \label{table:results-sd-spider} 
 \end{table*}
\end{center}

% --- Analysis ---
\vspace{-7mm}
Analyzing the individual questions, DocuT5-FK\textsuperscript{CD} does fix not only simple foreign key errors but also grounds natural language mentions to schema entities due to explicit context. 
By comparison, T5-Large\textsuperscript{CD} generally attempts to find the ``shortest path'' across tables.
In the first example in Figure \ref{fig:picard-vs-fk},  DocuT5-FK\textsuperscript{CD} fixes a Incorrect Table error type, because it receives explicit information that tables \textit{car\_data} and \textit{car\_names} are related. 
While in the second example, DocuT5-FK\textsuperscript{CD} has higher confidence in joining multiple tables together, and can better ground the implicit mention ``flights that arrive'' to the foreign key ``destairport'', which references table ``airports''.

Using the failure taxonomy developed in Section \ref{error-analysis}, we investigate how many Foreign Key failure types our model alleviates.
For a fair comparison, we inspect the questions predicted correctly by DocuT5-Large-FK\textsuperscript{CD}, which T5-Large\textsuperscript{CD} fails to infer accurately. 
Of this subset, our model can correctly predict 14 queries classified previously as Foreign Key failures. 
This represents 5\% of the total failed queries from the baseline T5-Large\textsuperscript{CD}.

Furthermore, we notice that a number of questions incorrectly predicted by T5-3b with constrained decoding were accurately predicted by our DocuT5-Base-FK\textsuperscript{CD} model, which shows that a careful input design can often outperform a significantly larger model. 
Compared to T5-3b, we improve 95 queries, from which 24 were direct Foreign Key mismatches. 
The rest of the improved queries were because our model combined tables better in aggregations to infer the correct SQL.

\subsubsection{Schema Descriptions}

% Examples from all three datasets (spider-dev, spider-dk and spider-syn in this order): https://docs.google.com/document/d/1DZ8vtIgOqmEwEtm6cUtlu3HMQ6peBZgVqvu_pvpu5mA/edit?usp=sharing .

% ----- Results ------

Table \ref{table:results-sd-spider} shows the schema description results where textual context is are explicitly encoded for tables and columns.
DocuT5-Large-SD\textsuperscript{CD} improves over T5-Large\textsuperscript{CD} by 2.3\% on EM and 1.8\% on EX on Spider dev. 
However, we see a much more significant relative improvement within the Spider-DK and Spider-SYN datasets.
DocuT5-Large-SD\textsuperscript{CD} improves over T5-Large\textsuperscript{CD} by 2.4\% EM in Spider-DK and 4.6\% EX on Spider-DK, and 2.8\% EM in Spider-DK and 3.7\% EX on Spider-SYN.
Considering a large amount of exact question-schema overlap in the original Spider dev set, Spider-DK and Spider-SYN are good means to assess generalization.
Adding schema descriptions is beneficial and should motivate other methods to leverage schema descriptions.
% Compared to comparable methods, specifically, RAT-SQL+B, which was previously the best-reported performing system, our method outperforms on Spider-DK by 7.1\% EM and 13.5\% on Spider-SYN.

% ----- Analysis ------

Using the failure taxonomy developed in Section \ref{error-analysis}, we investigate how many Foreign Key and Domain Knowledge failure types our model alleviates.
We find that DocuT5-Large-SD\textsuperscript{CD} improves a total of 47 queries: Foreign Key (13) and split across Domain Knowledge by Incorrect Aggregation (8), Incorrect Table (5), Incorrect Column (7), and Incorrect Value (11).

Figure \ref{fig:picard-vs-tableaug} shows specific examples where using schema descriptions helps for each failure category.
For example, DocuT5-Large-SD\textsuperscript{CD} can use implicit mentions of schema entities within the question to identify the correct column based on descriptions, i.e. identify both columns \textit{date\_effective\_from} and \textit{date\_effective\_to} when asked about ``effective date period''.
Extra column documentation on boolean columns is also beneficial in contexualising correct use cases, i.e. that ``left handed players" are players with column \textit{hand} set to ``L'' based on description ``a player is a right-handed player then \textit{hand} is \textit{R} otherwise \textit{L}''.

% Spider-DK
Looking at the Spider-DK dataset (Figure \ref{fig:picard-vs-tableaug-dk}), we find we can correctly predict queries belonging to the Logical Failures, such as questions asking for the ``youngest person'', which means the person with the minimum age or the maximum date of birth. 
The additional context through schema descriptions allows the model to reason more effectively when there is a question-table word mismatch.  
Furthermore, it is also clear that stating the column data type in the table documentation is advantageous; for example, "as a DateTime" results in the model getting less confused when identifying the correct logical operation.

% Spider-Syn
On the other hand, Figure \ref{fig:picard-vs-tableaug-dk} shows the results on the Spider-SYN dataset, we find additional context reduces some of the original hallucinations encountered. 
While we also observe the additional context makes the model more independent during inference.
Specifically, our model escapes some of the memorized patterns during training and adapts quickly to new domains, such as knowing that \textit{puppies} refers to the \textit{dogs} tables.

\section{Related Work}

% We outline the relevant text-to-SQL literature concerning table structure and schema descriptions.  

\textbf{Table Structure:} Recently released benchmarks, such as Spider \cite{yu2018spider} and WikiSQL \cite{zhong2017wikisql} make cross-domain text-to-SQL prediction more challenging than previous single domain datasets \cite{dahl1994atis, zelle1996geoquery}.
The cross-domain setting requires generalization to new database schemas and requires compositional SQL and identifying entity mentions in the natural language question. 

% In a cross-domain setting, generalization to new database schemas is critical because training and evaluation are performed on disjoint sets of databases. 
% Therefore, the model must recognize entity mentions in the natural language question and ground them to the schema entities, in addition to learning compositional SQL structures \cite{suhr2020exploring}.

Prior works has attempted to encode table structure by converting the schema into a directed graph \cite{bogin2019representing, guo2019towards} and adding global reasoning over the natural utterance through a question-contextualization \cite{wang2020rat}.
More recently, there have been a number of complex GNN-based approaches that have shown strong performance in the cross-domain SQL setting. 
Specifically, more advances schema grounded \cite{wang2022proton, cao2021lgesql}, including syntactic metadata \cite{hui2022s2sql}, and iteratively build a semantic enhanced schema-linking graph \cite{liu2022semantic}. 
Nevertheless, graph schema representations require complex model architectures with task-specific layers, and none of these methods enhance the schema-specific domain knowledge through the input.

Treating text-to-SQL as a machine translation problem, allows exploiting pre-trained language models that showed semantic knowledge understanding \cite{hwang2019comprehensive, li2020seqgensql}.
Most seq2seq SQL generators \cite{hwang2019comprehensive, lin2020bridging, scholak2021picard} only linearize the database schema by enumerating table and column names.
For example, PICARD \cite{scholak2021picard} and BRIDGE \cite{lin2020bridging} identified entity mentions within the natural utterance input through database look-ups and fuzzy string matching and appended them to the serialized database input.
Yet, as our behavior analysis identifies, complex table operations such as SQL join operations require richer table contextualization.   
UniSAr \cite{dou2022unisar} showed encoding foreign key relationships is beneficial, however, their results are worse on Spider and our method encodes cross-table relations in a much simpler manner.

\textbf{Schema Descriptions}: Relying solely on parametric memory learned during pre-training is not ideal, as revising the model's \textit{world knowledge} is non-trivial to do in practice. 
A hybrid model exploiting both parametric memory and a flexible external input allows easier adjustments \cite{lewis2020retrieval, shuster2021hallucination, izacard2022few}.
It can also reduce hallucinations and factual knowledge incorrectness, which are problems of generative models for knowledge intensive tasks \cite{roller2020chatbot}. 
% For example, even in a few-shot learning environment, ATLAS \cite{izacard2022few} showed impressive results using pre-trained retrieval augmented language models, outperforming non-retrieval models with 50x more parameters.    

In semantic parsing, input can be augmented by concatenating the top \textit{K} most similar natural utterance and logical form pairs. 
\citealp{gupta2021retronlu} augmented the inputs of a pre-trained language model with the top nearest neighbor semantic parses and \citealp{pasupat2021controllable} used custom retrieval index.

Another tangential work stream is pre-training seq2seq transformer-based models on
aligned tabular and textual data \cite{herzig2020tapas, yin2020tabert}.
However, the surrounding text is generally noisy and cannot inject much compositional bias.
We use a similar intuition: large pre-trained language models have been trained on free-form text, therefore high-quality table textual descriptions will provide better schema grounding in domain-specific knowledge.
% , by \textit{enriching} our model with schema-specific domain knowledge.
%%%%%%%%%%%%%%%%%%%%%%%%%%%%%%%%%%%%%%%%%%%%%%%%%%%%%%%%%%%%%%%%
%%%%%%%%%%%%%%%%%%%%%%%%%%%%%%%%%%%%%%%%%%%%%%%%%%%%%%%%%%%%%%%%

\section{Conclusion}

We introduce DocuT5, which allows diverse language model architectures to inject knowledge from external ``documentation'' to improve domain generalization. 
Through a newly developed failure taxonomy, we identify that current model errors are 19.6\% due to foreign key mistakes and 49.2\% due to a lack of domain knowledge. 
DocuT5 encodes knowledge from the table structure context of foreign keys and domain knowledge through contextualizing cross-table relations.
We show that both types of knowledge improve over state-of-the-art T5 with constrained decoding on SPIDER, and domain knowledge greatly helps on Spider-DK and Spider-SYN datasets.
Further analysis shows error reduction in foreign keys and domain knowledge failure categories and state-of-the-art performance over comparably pre-trained language SLQ generation models.

\section{Acknowledgements}
\label{sec:ack}

The authors would like  to acknowledge Ix Tech Global Ltd, specifically Tom Martin and Brie Read, for supporting this research.  
Additionally, this work is supported by the 2019 Bloomberg Data Science Research Grant and the Engineering and Physical Sciences Research Council grant EP/V025708/1. 

% Authors are encouraged to devote a section of their paper to concerns of the ethical impact of the work and to a discussion of broader impacts of the work, which will be taken into account in the review process. This discussion may extend into a 5th page (short papers) or 9th page (long papers).

% - no user was involved in the study
% - new datasets that people can use and build on
% - Potential risks of the work 
% - They have some extras if crow sourcing was used - but we didn't use them?

% \section*{Acknowledgements}

% Entries for the entire Anthology, followed by custom entries
\bibliography{acl_latex}

\begin{thebibliography}{28}
\expandafter\ifx\csname natexlab\endcsname\relax\def\natexlab#1{#1}\fi

\bibitem[{Bogin et~al.(2019)Bogin, Berant, and Gardner}]{bogin2019representing}
Ben Bogin, Jonathan Berant, and Matt Gardner. 2019.
\newblock Representing schema structure with graph neural networks for
  text-to-sql parsing.
\newblock In \emph{Proceedings of the 57th Annual Meeting of the Association
  for Computational Linguistics}, pages 4560--4565.

\bibitem[{Cao et~al.(2021)Cao, Chen, Chen, Zhao, Zhu, and Yu}]{cao2021lgesql}
Ruisheng Cao, Lu~Chen, Zhi Chen, Yanbin Zhao, Su~Zhu, and Kai Yu. 2021.
\newblock Lgesql: Line graph enhanced text-to-sql model with mixed local and
  non-local relations.
\newblock In \emph{Proceedings of the 59th Annual Meeting of the Association
  for Computational Linguistics and the 11th International Joint Conference on
  Natural Language Processing (Volume 1: Long Papers)}, pages 2541--2555.

\bibitem[{Dahl et~al.(1994)Dahl, Bates, Brown, Fisher, Hunicke-Smith, Pallett,
  Pao, Rudnicky, and Shriberg}]{dahl1994atis}
Deborah~A Dahl, Madeleine Bates, Michael~K Brown, William~M Fisher, Kate
  Hunicke-Smith, David~S Pallett, Christine Pao, Alexander Rudnicky, and
  Elizabeth Shriberg. 1994.
\newblock Expanding the scope of the atis task: The atis-3 corpus.
\newblock In \emph{Human Language Technology: Proceedings of a Workshop held at
  Plainsboro, New Jersey, March 8-11, 1994}.

\bibitem[{Dou et~al.(2022)Dou, Gao, Pan, Wang, Lou, Che, and
  Zhan}]{dou2022unisar}
Longxu Dou, Yan Gao, Mingyang Pan, Dingzirui Wang, Jian-Guang Lou, Wanxiang
  Che, and Dechen Zhan. 2022.
\newblock Unisar: A unified structure-aware autoregressive language model for
  text-to-sql.
\newblock \emph{arXiv preprint arXiv:2203.07781}.

\bibitem[{Gan et~al.(2021{\natexlab{a}})Gan, Chen, Huang, Purver, Woodward,
  Xie, and Huang}]{gan2021syn}
Yujian Gan, Xinyun Chen, Qiuping Huang, Matthew Purver, John~R Woodward, Jinxia
  Xie, and Pengsheng Huang. 2021{\natexlab{a}}.
\newblock Towards robustness of text-to-sql models against synonym
  substitution.
\newblock In \emph{Proceedings of the 59th Annual Meeting of the Association
  for Computational Linguistics and the 11th International Joint Conference on
  Natural Language Processing (Volume 1: Long Papers)}, pages 2505--2515.

\bibitem[{Gan et~al.(2021{\natexlab{b}})Gan, Chen, and Purver}]{gan2021dk}
Yujian Gan, Xinyun Chen, and Matthew Purver. 2021{\natexlab{b}}.
\newblock Exploring underexplored limitations of cross-domain text-to-sql
  generalization.
\newblock In \emph{Proceedings of the 2021 Conference on Empirical Methods in
  Natural Language Processing}, pages 8926--8931.

\bibitem[{Guo et~al.(2019)Guo, Zhan, Gao, Xiao, Lou, Liu, and
  Zhang}]{guo2019towards}
Jiaqi Guo, Zecheng Zhan, Yan Gao, Yan Xiao, Jian-Guang Lou, Ting Liu, and
  Dongmei Zhang. 2019.
\newblock Towards complex text-to-sql in cross-domain database with
  intermediate representation.
\newblock In \emph{Proceedings of the 57th Annual Meeting of the Association
  for Computational Linguistics}, pages 4524--4535.

\bibitem[{Gupta et~al.(2022)Gupta, Shrivastava, Sagar, Aghajanyan, and
  Savenkov}]{gupta2021retronlu}
Vivek Gupta, Akshat Shrivastava, Adithya Sagar, Armen Aghajanyan, and Denis
  Savenkov. 2022.
\newblock Retronlu: Retrieval augmented task-oriented semantic parsing.
\newblock In \emph{Proceedings of the 4th Workshop on NLP for Conversational
  AI}, pages 184--196.

\bibitem[{Herzig et~al.(2020)Herzig, Nowak, Mueller, Piccinno, and
  Eisenschlos}]{herzig2020tapas}
Jonathan Herzig, Pawel~Krzysztof Nowak, Thomas Mueller, Francesco Piccinno, and
  Julian Eisenschlos. 2020.
\newblock Tapas: Weakly supervised table parsing via pre-training.
\newblock In \emph{Proceedings of the 58th Annual Meeting of the Association
  for Computational Linguistics}, pages 4320--4333.

\bibitem[{Hui et~al.(2022)Hui, Geng, Wang, Qin, Li, Li, Sun, and
  Li}]{hui2022s2sql}
Binyuan Hui, Ruiying Geng, Lihan Wang, Bowen Qin, Yanyang Li, Bowen Li, Jian
  Sun, and Yongbin Li. 2022.
\newblock S2sql: Injecting syntax to question-schema interaction graph encoder
  for text-to-sql parsers.
\newblock In \emph{Findings of the Association for Computational Linguistics:
  ACL 2022}, pages 1254--1262.

\bibitem[{Hwang et~al.(2019)Hwang, Yim, Park, and Seo}]{hwang2019comprehensive}
Wonseok Hwang, Jinyeong Yim, Seunghyun Park, and Minjoon Seo. 2019.
\newblock A comprehensive exploration on wikisql with table-aware word
  contextualization.
\newblock \emph{arXiv preprint arXiv:1902.01069}.

\bibitem[{Izacard et~al.(2022)Izacard, Lewis, Lomeli, Hosseini, Petroni,
  Schick, Dwivedi-Yu, Joulin, Riedel, and Grave}]{izacard2022few}
Gautier Izacard, Patrick Lewis, Maria Lomeli, Lucas Hosseini, Fabio Petroni,
  Timo Schick, Jane Dwivedi-Yu, Armand Joulin, Sebastian Riedel, and Edouard
  Grave. 2022.
\newblock Few-shot learning with retrieval augmented language models.
\newblock \emph{arXiv preprint arXiv:2208.03299}.

\bibitem[{Lewis et~al.(2020)Lewis, Perez, Piktus, Petroni, Karpukhin, Goyal,
  K{\"u}ttler, Lewis, Yih, Rockt{\"a}schel et~al.}]{lewis2020retrieval}
Patrick Lewis, Ethan Perez, Aleksandra Piktus, Fabio Petroni, Vladimir
  Karpukhin, Naman Goyal, Heinrich K{\"u}ttler, Mike Lewis, Wen-tau Yih, Tim
  Rockt{\"a}schel, et~al. 2020.
\newblock Retrieval-augmented generation for knowledge-intensive nlp tasks.
\newblock \emph{Advances in Neural Information Processing Systems},
  33:9459--9474.

\bibitem[{Li et~al.(2020)Li, Keller, Butler, and Cer}]{li2020seqgensql}
Ning Li, Bethany Keller, Mark Butler, and Daniel Cer. 2020.
\newblock Seqgensql--a robust sequence generation model for structured query
  language.
\newblock \emph{arXiv preprint arXiv:2011.03836}.

\bibitem[{Lin et~al.(2020)Lin, Socher, and Xiong}]{lin2020bridging}
Xi~Victoria Lin, Richard Socher, and Caiming Xiong. 2020.
\newblock Bridging textual and tabular data for cross-domain text-to-sql
  semantic parsing.
\newblock In \emph{Findings of the Association for Computational Linguistics:
  EMNLP 2020}, pages 4870--4888.

\bibitem[{Liu et~al.(2022)Liu, Hu, Lin, and Wen}]{liu2022semantic}
Aiwei Liu, Xuming Hu, Li~Lin, and Lijie Wen. 2022.
\newblock Semantic enhanced text-to-sql parsing via iteratively learning schema
  linking graph.
\newblock In \emph{Proceedings of the 28th ACM SIGKDD Conference on Knowledge
  Discovery and Data Mining}, pages 1021--1030.

\bibitem[{Pasupat et~al.(2021)Pasupat, Zhang, and
  Guu}]{pasupat2021controllable}
Panupong Pasupat, Yuan Zhang, and Kelvin Guu. 2021.
\newblock Controllable semantic parsing via retrieval augmentation.
\newblock In \emph{Proceedings of the 2021 Conference on Empirical Methods in
  Natural Language Processing}, pages 7683--7698.

\bibitem[{Raffel et~al.(2020)Raffel, Shazeer, Roberts, Lee, Narang, Matena,
  Zhou, Li, and Liu}]{raffel2019t5}
Colin Raffel, Noam Shazeer, Adam Roberts, Katherine Lee, Sharan Narang, Michael
  Matena, Yanqi Zhou, Wei Li, and Peter~J Liu. 2020.
\newblock Exploring the limits of transfer learning with a unified text-to-text
  transformer.
\newblock \emph{Journal of Machine Learning Research}, 21:1--67.

\bibitem[{Roller et~al.(2021)Roller, Dinan, Goyal, Ju, Williamson, Liu, Xu,
  Ott, Smith, Boureau et~al.}]{roller2020chatbot}
Stephen Roller, Emily Dinan, Naman Goyal, Da~Ju, Mary Williamson, Yinhan Liu,
  Jing Xu, Myle Ott, Eric~Michael Smith, Y-Lan Boureau, et~al. 2021.
\newblock Recipes for building an open-domain chatbot.
\newblock In \emph{Proceedings of the 16th Conference of the European Chapter
  of the Association for Computational Linguistics: Main Volume}, pages
  300--325.

\bibitem[{Scholak et~al.(2021)Scholak, Schucher, and
  Bahdanau}]{scholak2021picard}
Torsten Scholak, Nathan Schucher, and Dzmitry Bahdanau. 2021.
\newblock Picard: Parsing incrementally for constrained auto-regressive
  decoding from language models.
\newblock In \emph{Proceedings of the 2021 Conference on Empirical Methods in
  Natural Language Processing}, pages 9895--9901.

\bibitem[{Shuster et~al.(2021)Shuster, Poff, Chen, Kiela, and
  Weston}]{shuster2021hallucination}
Kurt Shuster, Spencer Poff, Moya Chen, Douwe Kiela, and Jason Weston. 2021.
\newblock Retrieval augmentation reduces hallucination in conversation.
\newblock In \emph{Findings of the Association for Computational Linguistics:
  EMNLP 2021}, pages 3784--3803.

\bibitem[{Wang et~al.(2020)Wang, Shin, Liu, Polozov, and
  Richardson}]{wang2020rat}
Bailin Wang, Richard Shin, Xiaodong Liu, Oleksandr Polozov, and Matthew
  Richardson. 2020.
\newblock Rat-sql: Relation-aware schema encoding and linking for text-to-sql
  parsers.
\newblock In \emph{Proceedings of the 58th Annual Meeting of the Association
  for Computational Linguistics}, pages 7567--7578.

\bibitem[{Wang et~al.(2022)Wang, Qin, Hui, Li, Yang, Wang, Li, Sun, Huang, Si
  et~al.}]{wang2022proton}
Lihan Wang, Bowen Qin, Binyuan Hui, Bowen Li, Min Yang, Bailin Wang, Binhua Li,
  Jian Sun, Fei Huang, Luo Si, et~al. 2022.
\newblock Proton: Probing schema linking information from pre-trained language
  models for text-to-sql parsing.
\newblock In \emph{Proceedings of the 28th ACM SIGKDD Conference on Knowledge
  Discovery and Data Mining}, pages 1889--1898.

\bibitem[{Yin et~al.(2020)Yin, Neubig, Yih, and Riedel}]{yin2020tabert}
Pengcheng Yin, Graham Neubig, Wen-tau Yih, and Sebastian Riedel. 2020.
\newblock Tabert: Pretraining for joint understanding of textual and tabular
  data.
\newblock \emph{arXiv preprint arXiv:2005.08314}.

\bibitem[{Yu et~al.(2018)Yu, Zhang, Yang, Yasunaga, Wang, Li, Ma, Li, Yao,
  Roman et~al.}]{yu2018spider}
Tao Yu, Rui Zhang, Kai Yang, Michihiro Yasunaga, Dongxu Wang, Zifan Li, James
  Ma, Irene Li, Qingning Yao, Shanelle Roman, et~al. 2018.
\newblock Spider: A large-scale human-labeled dataset for complex and
  cross-domain semantic parsing and text-to-sql task.
\newblock In \emph{Proceedings of the 2018 Conference on Empirical Methods in
  Natural Language Processing}.

\bibitem[{Zelle and Mooney(1996)}]{zelle1996geoquery}
John~M Zelle and Raymond~J Mooney. 1996.
\newblock Learning to parse database queries using inductive logic programming.
\newblock In \emph{Proceedings of the national conference on artificial
  intelligence}, pages 1050--1055.

\bibitem[{Zhong et~al.(2020)Zhong, Yu, and Klein}]{zhong2020executionaccuracy}
Ruiqi Zhong, Tao Yu, and Dan Klein. 2020.
\newblock Semantic evaluation for text-to-sql with distilled test suites.
\newblock \emph{arXiv preprint arXiv:2010.02840}.

\bibitem[{Zhong et~al.(2017)Zhong, Xiong, and Socher}]{zhong2017wikisql}
Victor Zhong, Caiming Xiong, and Richard Socher. 2017.
\newblock Seq2sql: Generating structured queries from natural language using
  reinforcement learning.
\newblock In \emph{Proceedings of the 2020 Conference on Empirical Methods in
  Natural Language Processing (EMNLP)}.

\end{thebibliography}
\bibliographystyle{acl_natbib}

\appendix

\section{Appendix}
\label{sec:appendix}

\begin{center}
\begin{figure*}[h!]
\centering
    \includegraphics[scale=0.29]{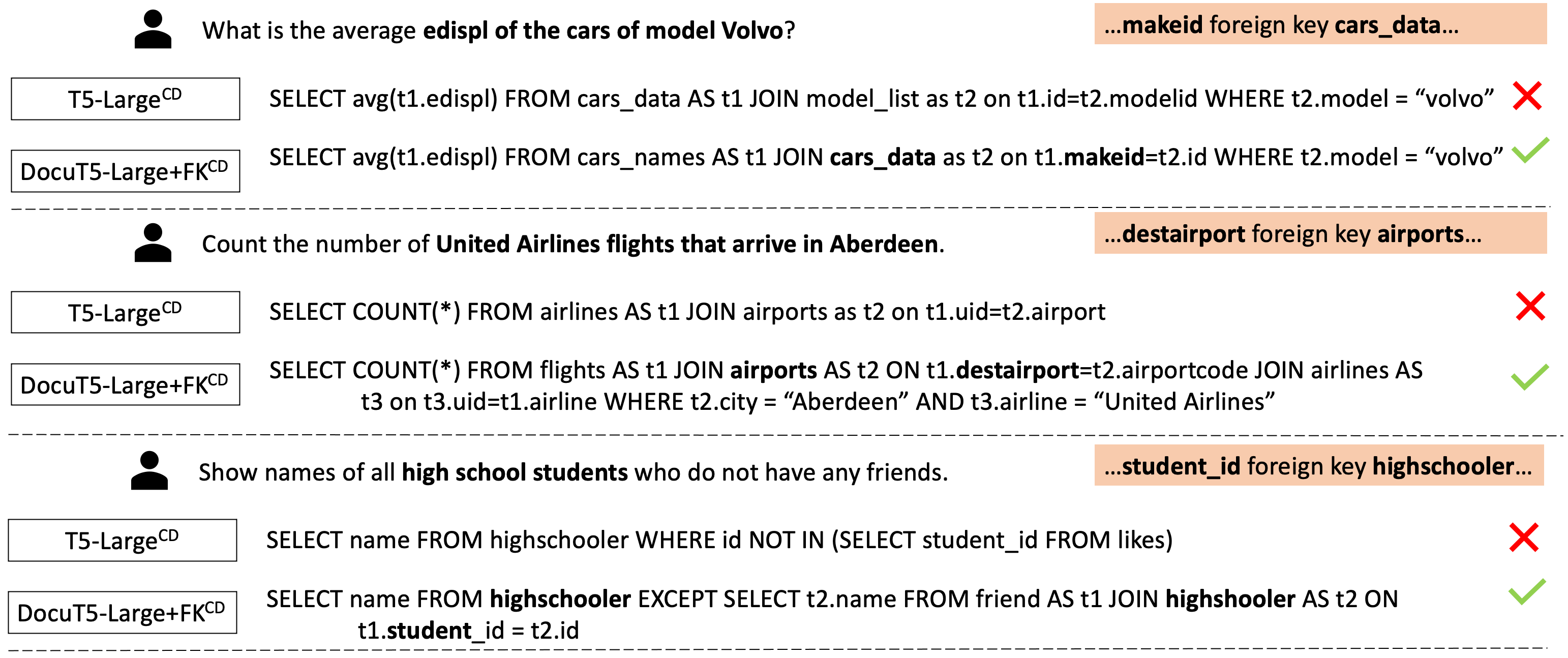}
    \caption{Spider dev: Text-to-SQL query comparing T5-Large\textsuperscript{CD} and DocuT5-Large+FK\textsuperscript{CD}, where regular T5 fails to infer that a JOIN was required or which foreign key to use. 
    The orange box is the augmented foreign key information  DocuT5-Large+FK\textsuperscript{CD} uses for the prediction.}
    \label{fig:picard-vs-fk}
\end{figure*}
\end{center}

\begin{center}
\begin{figure*}[h!]
\centering
    \includegraphics[scale=0.32]{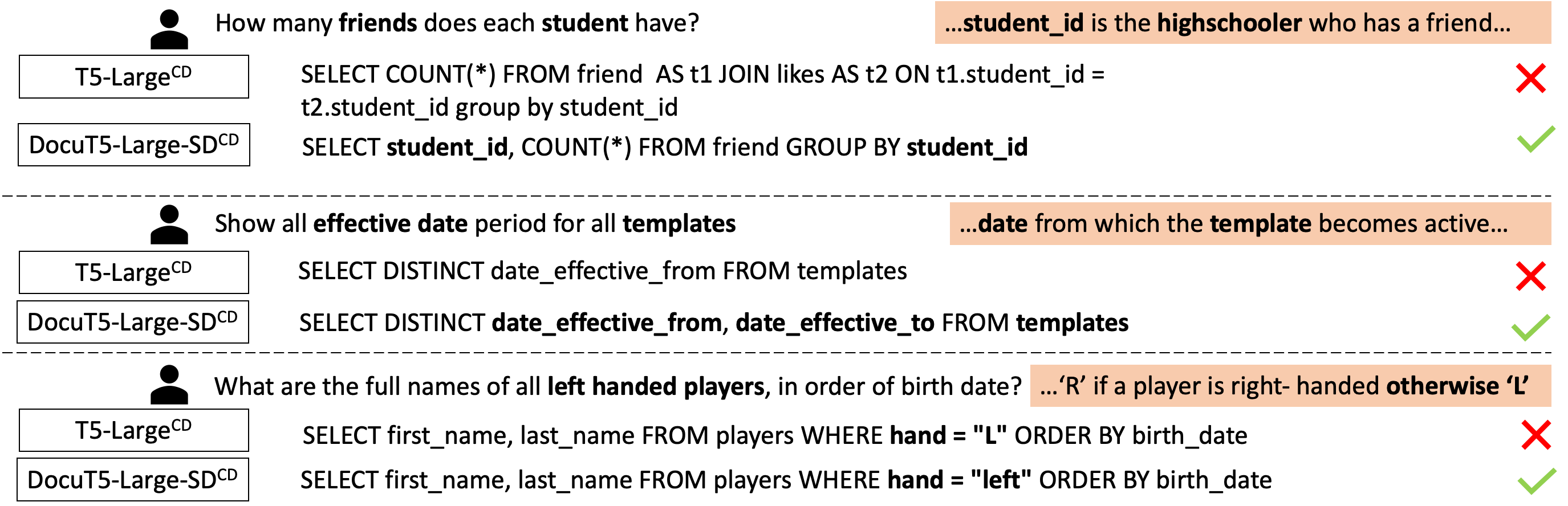}
    \caption{Spider dev: Text-to-SQL examples comparing T5-Large\textsuperscript{CD} and DocuT5-Large+SD\textsuperscript{CD}, where regular T5 fails due to lack of domain knowledge. 
    The orange box is the augmented table and column information DocuT5-Large+SD\textsuperscript{CD} uses for the prediction.}
    \label{fig:picard-vs-tableaug}
\end{figure*}
\end{center}

\begin{center}
\begin{figure*}[h!]
\centering
    \includegraphics[scale=0.32]{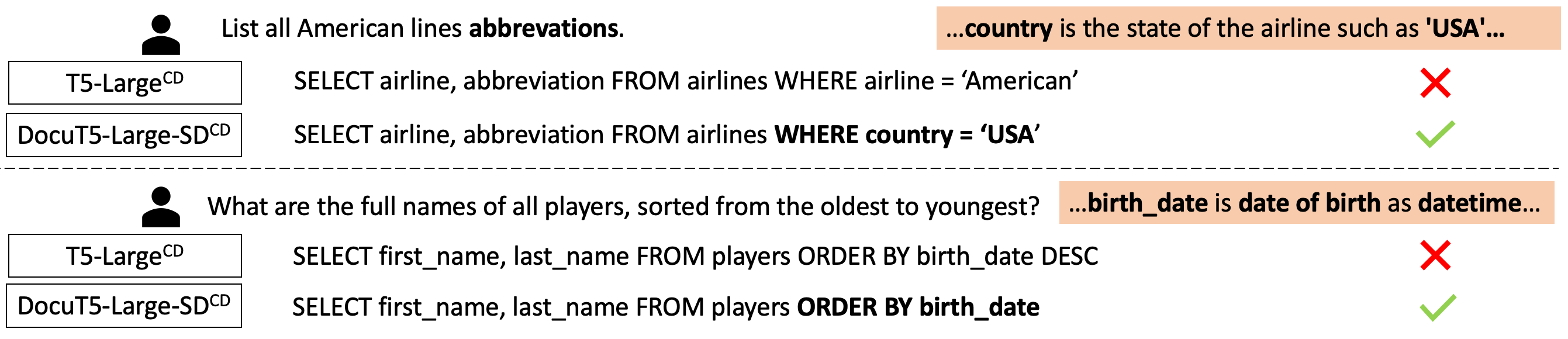}
    \caption{Spider-DK: Text-to-SQL examples comparing T5-Large\textsuperscript{CD} and DocuT5-Large+SD\textsuperscript{CD}, where regular T5 fails due to lack of domain knowledge. 
    The orange box is the augmented table and column information DocuT5-Large+SD\textsuperscript{CD} uses for the prediction.}
    \label{fig:picard-vs-tableaug-dk}
\end{figure*}
\end{center}

\begin{center}
\begin{figure*}[h!]
\centering
    \includegraphics[scale=0.32]{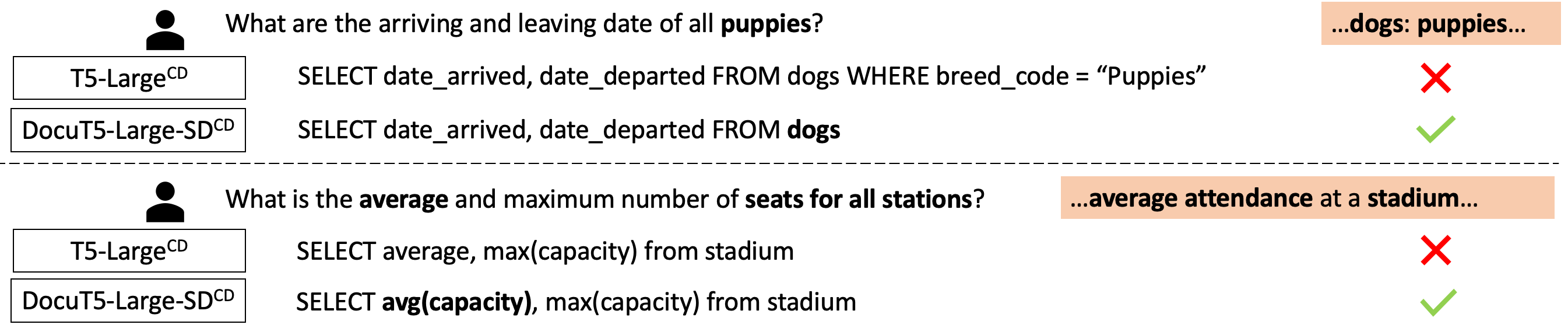}
    \caption{Spider-SYN: Text-to-SQL examples comparing T5-Large\textsuperscript{CD} and DocuT5-Large+SD\textsuperscript{CD}, where regular T5 fails due to lack of domain knowledge. 
    The orange box is the augmented table and column information DocuT5-Large+SD\textsuperscript{CD} uses for the prediction.}
    \label{fig:picard-vs-tableaug-syn}
\end{figure*}
\end{center}

% This is an appendix.

\end{document}